\documentclass[letterpaper, 10 pt, conference]{ieeeconf}  

\IEEEoverridecommandlockouts 

\usepackage{tabularx}
\usepackage{array}
\usepackage{multirow}
\usepackage{booktabs}
\usepackage{comment}
\usepackage{siunitx}
\usepackage{caption}
\usepackage{makecell}
\usepackage{xcolor}
\usepackage{textcase} 
\usepackage{cite}
\usepackage{amsmath,amssymb,amsfonts}
\usepackage{graphicx}
\usepackage{textcomp}
\usepackage{lettrine}   
\usepackage{relsize}
\usepackage{todonotes}
\usepackage{cuted}
\usepackage{algorithm}
\usepackage{float}
\usepackage{subcaption}

\usepackage{tabularx}
\usepackage{booktabs}
\usepackage{multirow}
\usepackage{array}
\usepackage{calc}
\usepackage{hyperref}
\usepackage{textcomp}

\usepackage[noend]{algpseudocode}

\definecolor{mygreen}{RGB}{34,139,34}
\title{\LARGE \bf
Push-Wiper: Toward General-Purpose Robotic Cleaning across Varied Stains and Surfaces with Segmented Pushing Trajectories}

\author{
Renhao Lu$^{1,2,\dagger}$,
Mingxin Wang$^{1,\dagger}$,
Chenyang Cao$^{3}$,
Yang Yang$^{1}$,
Guoping Pan$^{1,2}$,\\
Kangkang Dong$^{1}$,
Yi Cheng$^{2}$,
Houde Liu$^{1,2,*}$%
\thanks{$^{1}$Tsinghua Shenzhen International Graduate School, Tsinghua University, Shenzhen 518055, China.}%
\thanks{$^{2}$Z-Lab, Zerith Robotics, Shenzhen 518055, China.}%
\thanks{$^{3}$University of Toronto, Toronto M5S 1A1, Canada.}%
\thanks{$^{*}$Corresponding author: Houde Liu (liu.hd@sz.tsinghua.edu.cn).}%
\thanks{This work was supported by the Shenzhen Science and Technology Program 
(Grant No. RCJC20210706091946001) and the Shenzhen Science and Technology Program 
(Grant No. ZDCY20250901104207008). ($^{\dagger}$ indicates equal contribution).}
}

\begin{document}

\maketitle
\begin{strip}
\begin{minipage}{\textwidth}\centering
\vspace{-30pt}
\includegraphics[width=1\textwidth]{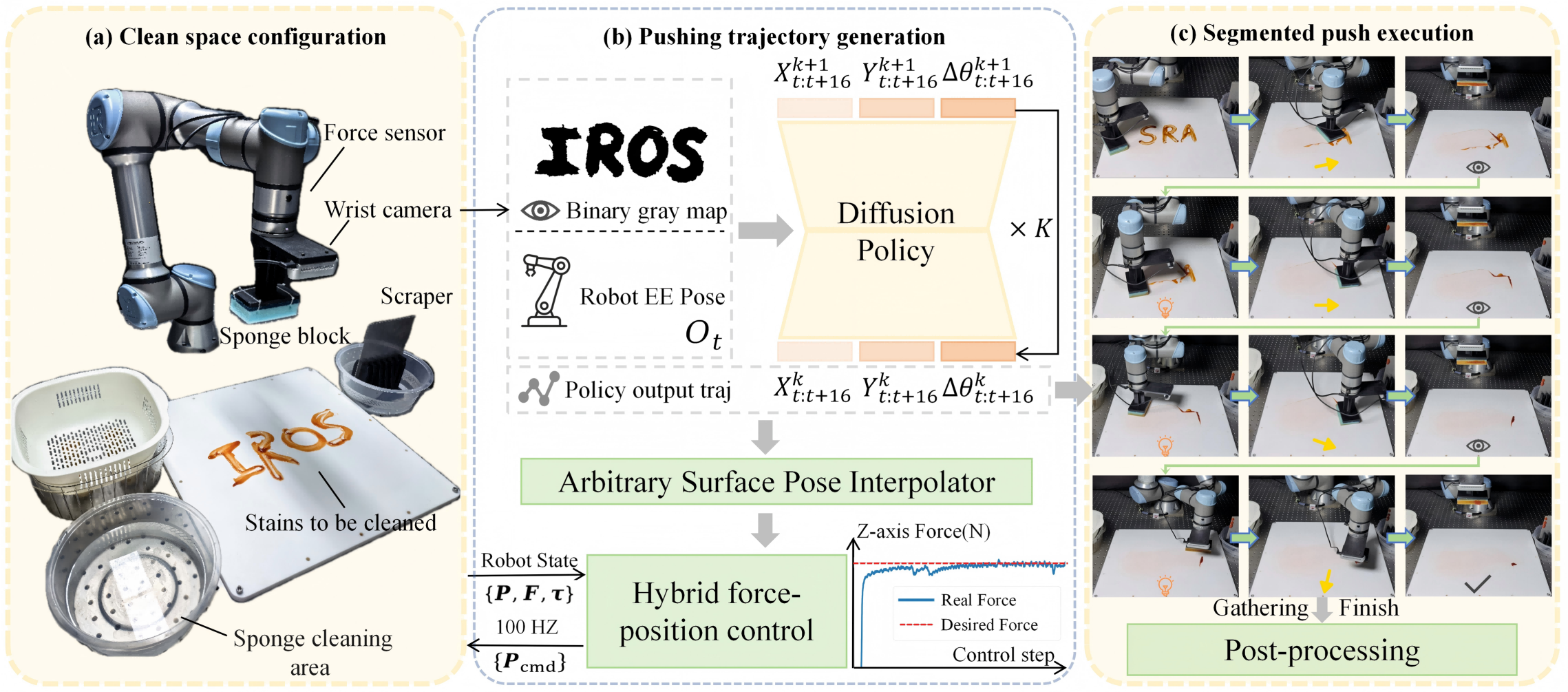}
\captionof{figure}{\textbf{Overview of the Push-Wiper framework.}
(a) Experimental setup and tools configuration.
(b) Trajectory generation: From one observation, Push-Wiper uses Diffusion Policy to produce 3D pushing actions ($x, y, \Delta\theta$), which are converted into a smooth executable trajectory by our Arbitrary Surface Pose Interpolator and  Hybrid force–position controller.
(c) Through iterative \emph{Perception–Planning–Execution} cycles, the robot gathers stains into a small region, after which Post-processing removes the residues to complete cleaning.}
\label{fig:framework}
\end{minipage}
\end{strip}
\thispagestyle{empty}
\pagestyle{empty}

\begin{abstract}
Viscous stains, characterized by high viscosity and complex rheological properties, remain a major challenge for robotic surface cleaning. Conventional wiping often spreads the stain and scrubbing provides stronger friction but risks damaging the surface. In this paper, we propose Push-Wiper, a framework that reformulates viscous stain cleaning as an aggregation problem. Push-Wiper employs a sponge to progressively gather stains through segmented pushing trajectories, followed by a post-processing phase that detaches the aggregated material and enables sponge self-cleaning. We adopt a stepwise strategy for stain gathering and leverage Diffusion Policy to generate adaptive pushing action sequences and execute them via our Arbitrary Surface Pose Interpolator (ASPI) together with a Hybrid force–position controller, allowing the method to generalize to stains with diverse spatial distributions. Push-Wiper achieves a cleaning score (CS), defined as the percentage of stain area removed, up to 130\% higher than baseline methods. Without additional training, Push-Wiper also transfers in a zero-shot manner to solid residues, liquid spills, unseen viscous stains and curved surfaces with varying geometries. Our experiments demonstrate the cleaning effectiveness of Push-Wiper and its strong generalization ability. The project website is available at \url{https://push-wiper.github.io/}. 
\end{abstract}

\section{Introduction}
Cleaning robots are now widely deployed in domestic, commercial, and industrial settings, with rapid advances in capability and scope in recent years \cite{megalingam2025cleaning}.
Similarly, the task of wiping has been automated for cleaning regular surfaces like floors \cite{bordoloi2017floor}, windows\cite{li2021survey}, and tables \cite{harmatz2024hybrid}. Although conventional robotic cleaning methods, such as sweeping and wiping, can effectively address solid residues and liquid spills in tabletop tasks, viscous stains continue to pose a significant challenge \cite{kowalewski2025sccrub}.

Existing in a semi-solid state, viscous stains exhibit high viscosity and complex rheological properties \cite{landel2021fluid}. Consequently, they are hard to remove and often smear or spread, worsening contamination. Effective cleaning therefore requires applying a controlled normal force and generating sufficient tangential friction to shear the material off the surface \cite{kim2019control}. Unlike dishwashing, where abundant rinsing is feasible \cite{wakabayashi2024behavioral}, tabletop settings restrict liquid use, further complicating cleaning.

Wiping and scrubbing are two common motions in robotic surface cleaning \cite{kowalewski2025sccrub}. Scrubbing, which employs tools like brushes in reciprocating motions, is effective for removing viscous stains. However, it poses a higher risk of surface abrasion and thus demands precise force control. Wiping, on the other hand, is highly effective for absorbable contaminants like liquid spills, but it is prone to causing secondary contamination by spreading viscous substances. Additionally, in real-world scenarios, tabletop stains are not uniform and can manifest in various forms, such as solid debris, liquid spills, and viscous substances \cite{johansson2007handbook}. Therefore, a cleaning robot can hardly rely on a single strategy or tool for effective cleaning. This often requires the combination of different approaches \cite{yin2020table, Lew2023}, which in turn leads to greater complexity in both design and implementation.


In this work, we propose Push-Wiper, a learning-based framework specifically designed to tackle the complex rheological properties of viscous stains (Fig.~\ref{fig:framework}). Push-Wiper fundamentally reformulates viscous stain cleaning as a topological aggregation problem. To bypass the intractable analytical modeling of fluid deformation under contact, we propose an ``aggregate-then-finish'' paradigm. In the gathering phase, Push-Wiper operates episodically. It utilizes Diffusion Policy \cite{chi2023diffusion} as a low-frequency action generator, inferring segmented low-dimensional pushing actions strictly from abstract binary observations. This deliberate abstraction forces the policy to learn multimodal aggregation behaviors independent of complex visual textures or 3D surface geometries. To execute these local aggregation steps, Push-Wiper employs an Arbitrary Surface Pose Interpolator (ASPI) combined with a hybrid force--position controller. This architecture strictly decouples 2D topological planning from 3D geometric execution, directly mapping the low-dimensional action sequence to a high-dimensional end-effector pose trajectory compatible with arbitrary surfaces. Finally, the post-processing phase introduces motion primitives to detach aggregated residue and self-clean the sponge for reuse. This decoupled design improves cleaning performance and enables robust zero-shot generalization to unseen curved surfaces and diverse stain categories without additional data or retraining.

In summary, our main contributions are as follows:

\begin{itemize}
\item \textbf{A Novel Paradigm for Viscous Stain Cleaning:} We cast viscous stain removal as a state-aggregation problem and propose an ``aggregate-then-finish'' strategy that overcomes the limitations of conventional wiping and scrubbing.

\item \textbf{A Decoupled Visuomotor Architecture:} We propose Push-Wiper, which couples a low-frequency Diffusion Policy for 3D action generation with ASPI and hybrid force--position control for 6D trajectory. This decoupling isolates the policy from 3D surface variations, reducing problem complexity.

\item \textbf{Strong Zero-Shot Generalization:} We demonstrate that, by abstracting the task into geometric topologies rather than complex visual textures or 3D surface geometries, our system generalizes zero-shot to unseen curved surfaces, solid residues, liquid spills, and novel viscous stains. Experiments show Push-Wiper achieves near-complete cleaning and improves cleaning score by up to 130\% over baselines.

\end{itemize}


\section{Related Work }

In this section, we review existing methods for robotic surface cleaning. Based on their approach to generating action trajectories, these methods can be broadly categorized into two groups: classical and learning-based methods.


\textbf{Classical methods.} Classical robotic surface cleaning methods typically rely on predefined trajectories executed through position, force, or compliant control. Hess et al.~\cite{hess2012null} and Wang et al.~\cite{Wang2025Hierarchically} perform coverage path planning and track the planned wiping motions using position control. To handle physical interaction with the environment, Ortenzi et al.~\cite{ortenzi2014experimental} incorporate contact constraints for robotic whiteboard wiping, while Leidner et al.~\cite{leidner2016knowledge} achieve compliant surface cleaning through impedance control~\cite{hogan1984impedance}. For viscous stains, Harmatz et al.~\cite{harmatz2024hybrid} and Kowalewski et al.~\cite{kowalewski2025sccrub} adopt hybrid force–position control with soft robotic arms. However, they still have some limitations: the former requires human intervention, while the latter still leaves substantial residues.

\textbf{Learning-based methods.} Learning-based cleaning methods mainly follow two paradigms: reinforcement learning (RL) and imitation learning (IL). RL formulates cleaning as reward-driven policy learning. Lew et al.~\cite{Lew2023} design rewards for liquid-spill cleaning to learn high-level wiping strategies, but their policy relies only on position control despite the importance of force feedback. Mart{'\i}n-Mart{'\i}n et al.~\cite{martin2019variable} combine PPO with variable impedance control for wiping, yet may suffer performance degradation during sim-to-real transfer. IL learns cleaning strategies from expert demonstrations. Tsuji et al.~\cite{tsuji2024adaptive} incorporate real-time force-torque feedback and pre-trained object representations into IL, enabling generalization to plane-height variations. Oishi et al.~\cite{oishi2025imitation} propose a motion generation model that modulates velocity and contact force during command-conditioned whiteboard wiping. Recently, diffusion models have been increasingly applied to robotics~\cite{song2025survey}, with several works~\cite{zhou2025admittance,he2025foar,xue2025reactive} integrating force information into diffusion-based wiping policies. However, these methods mainly address simple mark-wiping tasks on a single surface type, such as whiteboards, and do not tackle viscous stain cleaning in diverse environments.

\textbf{Summary.} Classical methods are stable for predefined surface cleaning tasks but rely on precise models and hand-tuned parameters, limiting their portability across platforms and environments. Learning-based methods improve adaptability within specific scenarios, yet generalization to new conditions often requires redesign or retraining. Currently, there is no single strategy that can effectively handle multiple types of stains on arbitrary curved surfaces. To the best of our knowledge, no learning-based framework has been specifically developed for the challenging task of cleaning viscous stains.

\section{Push-Wiper Framework}
\subsection{Problem decomposition} 
The objective of our work is to completely remove viscous stains from the surface. We simplify the surface $D$ into a discrete $M \times N$ grid, where the state of each cell is denoted by $D_{i,j} \in \{0, 1\}$. $0$ and $1$ represent a dirty state and a clean state, respectively. Our objective is formulated as follows:
\begin{equation}
\begin{aligned}
\max_{\{D_{i,j}\}} \quad & \sum_{i=1}^{M}\sum_{j=1}^{N} D_{i,j} \\
\text{s.t.}\quad & D_{i,j} \in \{0,1\}.
\end{aligned}
\end{equation}
For broad compatibility across robotic systems, our framework requires only basic hardware: a manipulator with a wrist camera and a simple tool, such as a sponge.

Directly wiping viscous stains often exacerbates contamination due to their complex rheology. Inspired by human strategies, Push-Wiper (Fig. \ref{fig:framework}) decouples the task into a \textit{gathering phase} and a \textit{post-processing phase}. Let $\mathcal{S}_t$ denote the set of stain pixels at step $t$. We explicitly formulate stain aggregation as minimizing its maximum spatial diameter $\mathcal{D}(\mathcal{S}_t)$ and connectivity fragmentation $K(\mathcal{S}_t)$ via segmented pushing trajectories $\tau$:
\begin{equation}
\min_{\tau} \quad \mathcal{D}(\mathcal{S}_{t+1}) + \lambda K(\mathcal{S}_{t+1}),
\label{eq:aggregation}
\end{equation}
where $\lambda$ balances the spatial spread and the number of disconnected components. Analytically predicting such fluid deformation under contact requires precise prior knowledge of hidden rheological parameters, making classical planning intractable \cite{billard2019trends, xian2023fluidlab}. This physically motivates our data-driven approach. Using binary maps and robot states as observations, the gathering phase progressively contracts the stain to a compact state (i.e. $\mathcal{D} < \epsilon, K \to 1$). Finally, the post-processing phase executes predefined primitives to remove the gathered residue, achieving complete cleaning.

\begin{algorithm}[!b]
\caption{ArbitrarySurfacePoseInterpolator($\mathbf{A}_t,\,\mathbf{p}_{cap},\,f$)}
\label{Euclid}
\begin{algorithmic}[1]
\State $\mathbf{P}_{seq} \gets \{\}$
\For{$\mathbf{a} \in \mathbf{A}_t$}
  \State $x_{m},\,y_{m},\,\Delta\theta \gets PointInSurface(\mathbf{a},\mathbf{p}_{cap})$%
  \State $z_{m},\mathbf{n}_{m} \gets GetSurfaceNormal(x_{m},y_{m})$
  \State $\mathbf{p}_{base} \gets GeneratePose(x_{m},y_{m},z_{m},-\mathbf{n}_{m},\Delta\theta)$
  \State $\mathbf{P}_{seq}\gets\mathbf{P}_{seq}\cup\{\mathbf{p}_{base}\}$
\EndFor
\State $\mathbf{t}_{seq},\mathbf{\phi}_{seq} \gets \mathbf{P}_{seq}$
\State $\mathbf{t}_{traj} \gets BslineAndTrapVel(\mathbf{t}_{seq},\,f)$
\State $\mathbf{\phi}_{traj} \gets Slerp(\mathbf{\phi}_{seq},\,f)$
\State $\mathbf{P}_{traj} \gets Hstack(\mathbf{t}_{traj},\,\mathbf{\phi}_{traj})$
\State Return $\mathbf{P}_{traj}$

\end{algorithmic}
\end{algorithm}

\subsection{Gathering Phase}

\textbf{Action generation.} To align our data-driven approach with the aggregation objective in Eq. \eqref{eq:aggregation}, we deliberately abstract the visual observation into a texture-less binary stain map $\mathbf{M}_t$ ($\text{stain}=0, \text{clean}=1$). This geometric abstraction eliminates the policy’s reliance on visual appearance, forcing the robot to learn pushing strategies purely from the stain’s spatial topology.
Because a scattered stain distribution can be aggregated from multiple valid directions, we adopt Diffusion Policy (DP) \cite{chi2023diffusion} to capture this inherent multimodal distribution of expert demonstrations.

Standard DP operates in a high-frequency receding-horizon manner with continuous visual feedback, which cannot be reliably maintained under our minimal sensing setting: the wrist camera loses view during contact, and an external camera view is often occluded by the arm. Conversely, planning the entire gathering phase from a single initial observation is also ineffective, as the severe redistribution of viscous stains during pushing quickly invalidates the initial plan. To explicitly address these dual challenges and encode the aggregation behavior, we curate our expert demonstrations as discrete, segmented pushing strokes rather than continuous, end-to-end cleaning episodes. We view each pushing trajectory as a local optimization step that aims to decrease the aggregation objective
$J_t=\mathcal{D}(\mathcal{S}_t)+\lambda K(\mathcal{S}_t)$ in Eq.~\eqref{eq:aggregation}. Consequently, we deploy DP as a low-frequency, macro-level planner: re-perceive, plan, execute open-loop, and re-perceive. 

The iterative gathering process is shown in Fig. \ref{fig:framework}(b) and (c). At the beginning of step $t$, the observation $\mathbf{O}_t = \{\mathbf{M}_t, \mathbf{p}_{cap}\}$ (where $\mathbf{p}_{cap}$ is the fixed capture pose) is fed into the policy $\pi$ to generate an $n$-step action sequence $\mathbf{A}_{t}$. Because of our segmented training paradigm, $\mathbf{A}_{t}$ does not merely represent a short-term receding horizon, but strictly encodes a complete pushing trajectory. This temporal discretization transforms DP into a local aggregation operator: at each macro-step, it infers a full stroke to progressively reduce the stain's spatial diameter.

Crucially, we restrict the action space to $a = (x_b, y_b, \Delta\theta)$, representing 2D planar translations and yaw rotations relative to $\mathbf{p}_{cap}$. Rather than outputting arbitrary 6D trajectories, this low-dimensional space strictly regularizes the policy to generate planar aggregating motions that push the boundaries of $\mathbf{M}_t$ inward. By decoupling this 2D topological planning from the 3D surface geometry, ASPI can independently reconstruct the 6-DoF end-effector pose. This design inherently grounds the learned actions in our aggregation formulation, drastically reducing the search space and enabling zero-shot generalization across unseen curved surfaces.

\textbf{Trajectory generation.} Algorithm 1 outlines ASPI. While mapping low-dimensional trajectories onto 3D surfaces is an effective approach for robotic operations\cite{2019Song}, we uniquely deploy it here to explicitly decouple 2D topological inference from 3D geometric execution. For each 3D planar action $\mathbf{a} \in \mathbf{A}_{t}$, ASPI projects it onto the 3D surface to extract the corresponding point $(x_{m},y_{m},z_{m})$ and normal vector $\mathbf{n}_{m}$ (lines 1-6). To ensure effective gathering of viscous stains and maintain stable physical contact, we enforce a strict geometric pose constraint on the end-effector:  the z-axis of the TCP is strictly aligned with $-\mathbf{n}_{m}$, and the model predicts an additional yaw increment $\Delta\theta$, which is superimposed on the reference pose $p_{cap}$. Furthermore, because our policy outputs sparse macro-waypoints, direct execution would induce acceleration transients that destabilize the subsequent hybrid force-position controller. Thus, ASPI applies B-spline fitting, trapezoidal velocity time parameterization, and spherical linear interpolation to synthesize a 6D trajectory $\mathbf{P}_{traj}$ at $f$ Hz (lines 7-10). Crucially, rather than requiring the intractable collection of expert demonstrations across diverse curved geometries, this deterministic 2D-to-3D bridge explicitly shields the policy from the burden of modeling complex 3D surface variations. By learning topological aggregation exclusively from planar data, we radically reduce the problem complexity, which inherently enables our zero-shot generalization across unseen curved surfaces.

\textbf{Trajectory execution.}
To push viscous stains while executing $\mathbf{P}_{traj}$, Push-Wiper maintains a prescribed contact force to keep the sponge in close contact with the surface.
In our setting, successful execution primarily depends on maintaining sufficient normal contact during motion. We therefore handle force feedback at the execution layer rather than using it as a policy input. 
We therefore keep force feedback in the execution layer and adopt a hybrid force--position controller that decouples force regulation from motion tracking.
Specifically, Push-Wiper uses an admittance controller~\cite{ott2010unified} to regulate the normal contact force to a constant setpoint $F_{z}^{des}$.
Since each pose in $\mathbf{P}_{traj}$ aligns the TCP $z$-axis with the surface normal, this is equivalent to regulating the TCP $z$-axis force component to $F_{z}^{des}$. The admittance controller is as follows: 
\begin{equation}\label{eq:admittance}
m\,\ddot{\delta z}+b\,\dot{\delta z}+k\,\delta z = F_{z}^{des} - F_{z},
\end{equation}
where $m,b,k$ denote the inertia, damping and stiffness matrices. $F_{z}$ is the feedback of the force sensor and $\delta z$ is the correction value of the admittance controller. The remaining translational axes and end-effector orientation follow $\mathbf{P}_{traj}$ under position control. In the TCP frame, the x--y tracking error $\Delta\mathbf{s}^{E}_{\mathrm{track}}$ and admittance increment $\Delta\mathbf{s}^{E}_{\mathrm{adm}}$ are:
\begin{equation}\label{eq:s-track-adm}
\Delta\mathbf{s}^{E}_{\mathrm{track}}
= \Big[(\mathbf{R}^{B}_{E})^{\top}\!\big(\mathbf{s}^{B}_{\mathrm{traj}}-\mathbf{s}^{B}_{E}\big)\Big]_{x,y},\,
\Delta\mathbf{s}^{E}_{\mathrm{adm}}=
\begin{bmatrix}0\\0\\\delta z\end{bmatrix},
\end{equation}
where $\mathbf{R}^{B}_{E}$ is the rotation matrix from the TCP frame to the base frame. $\mathbf{s}^{B}_{\mathrm{traj}}$ and $\mathbf{s}^{B}_{\mathrm{E}}$ denote the desired trajectory position and the measured end-effector position in the base frame, respectively. The final pose command sent to the manipulator is:
\begin{equation}\label{eq:s-command}
\mathbf{s}^{B}_{\mathrm{cmd}}
= \mathbf{s}^{B}_{E}
+ \mathbf{R}^{B}_{E}\!\left(\Delta\mathbf{s}^{E}_{\mathrm{track}}+\Delta\mathbf{s}^{E}_{\mathrm{adm}}\right),\,
\mathbf{R}^{B}_{\mathrm{cmd}}=\mathbf{R}^{B}_{\mathrm{traj}},
\end{equation}
where $\mathbf{R}^{B}_{\mathrm{traj}}$ is the orientation in expected trajectory point.

To reduce residual stains, after each pushing trajectory the manipulator performs a predefined scraping motion to clean the sponge. During gathering, it repeats pushing until the stain area drops below $q_{th}$.

\subsection{Post-processing Phase}
\begin{figure}[t]   
    \centering 
    \vspace{1ex}
    \includegraphics[width=1\linewidth]{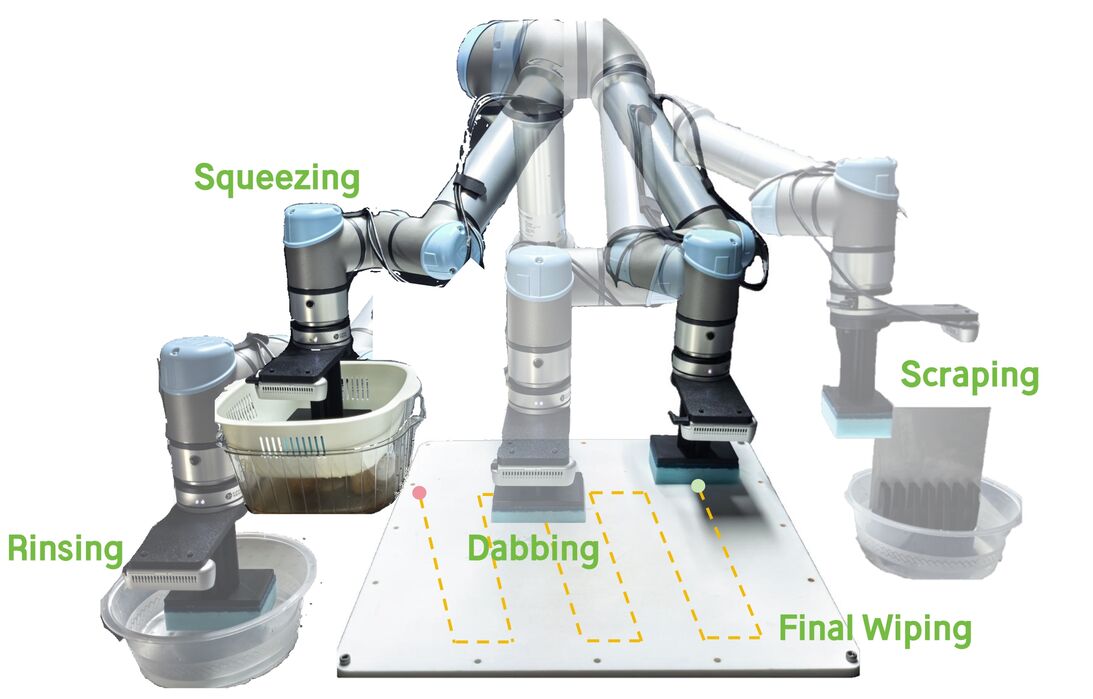}  
    \caption{\textbf{Five motion primitives in the post-processing phase.} The yellow dashed line indicates the full-cover path in the \textit{final wiping}.}
    \label{fig:post} 
\end{figure}

The post-processing phase removes the gathered residue and completes the cleaning process.
As shown in Fig.~\ref{fig:post}, after the gathering phase aggregates the stain into a compact region, the robot executes a small set of predefined motion primitives with the same hybrid force--position controller.
These primitives include dabbing to lift the residue, scraping, rinsing, and squeezing to self-clean the sponge and regulate moisture, followed by a final full-coverage wipe. This phase improves cleaning completeness and supports repeated sponge use within the framework.

\section{Experiments}

Our experiments include: (i) comparative evaluation of Push-Wiper (without post-processing) against two baselines on two representative viscous stains, (ii) validation of its generalization on unseen objects and arbitrary curved surfaces, and (iii) evaluation of the effectiveness of post-processing.

\subsection{Experimental Setup}

\textbf{Platform.} As shown in Fig.~\ref{fig:framework}(a), we use a UR7e robot with a KWR75B six-axis force/torque sensor for hybrid force--position control and an Intel RealSense D435 wrist camera for capturing stain images. An \SI{11}{\centi\meter} $\times$ \SI{7}{\centi\meter} sponge block is mounted via a 3D-printed adapter, requiring no specialized mechanisms. All devices are connected to a workstation with an Intel Core i7-14700F CPU and NVIDIA RTX 4060Ti GPU for data collection and policy evaluation.

\textbf{Tasks.} We categorize viscous stains into two types:
\begin{itemize}
    \item \textbf{Simple} type: the surface contains only one connected stain region, and the stain pixels account for less than \SI{20}{\percent} of the entire image, such as simple shapes resembling numbers or letters.
    \item \textbf{Complex} type: the surface contains two or more connected stain regions, or the stain pixels account for more than \SI{20}{\percent} of the image, such as multiple numbers distributed in a scattered manner.
\end{itemize}

\textbf{Metrics.} We evaluate our system on two representative viscous stains with distinct physical properties: ketchup and peanut butter. Ketchup exhibits an apparent viscosity of \SIrange{1000}{1500}{mPa\cdot s} at a shear rate of \SI{10}{s^{-1}} \cite{kumbar2019rheological}, whereas peanut butter reaches \SI{29452}{mPa\cdot s} under the same condition \cite{citerne2001rheological}.
This near order-of-magnitude gap implies much higher adhesion for peanut butter, making it more challenging to clean.

To quantitatively compare the cleaning performance under these challenging conditions, we define the Cleaning Score (CS).
Specifically, we extract a denoised stain mask using an area-based detector that fuses HSV/Lab/grayscale cues with simple morphology and connected-component filtering.
Let \(N_{\text{before}}\) and \(N_{\text{after}}\) denote the total number of stain pixels before and after each cleaning, respectively. The Cleaning Score (CS) is then defined as the percentage reduction in stain pixels after cleaning:
\begin{equation}
\text{CS} = \left( 1 - \frac{N_{\text{after}}}{N_{\text{before}}} \right) \times 100
\end{equation}

\subsection{Baselines and Implementation}

\textbf{Baselines.} 
To isolate the benefit of the proposed aggregation-first paradigm under a controlled and reproducible setup, we implement two strong baseline strategies within the same perception and execution stack. This design minimizes confounding factors from implementation differences, while covering two common alternatives: coverage-style wiping and one-shot pushing.

\begin{figure}[htbp]
    \centering
    \vspace{1ex}
    \includegraphics[width=1\linewidth]{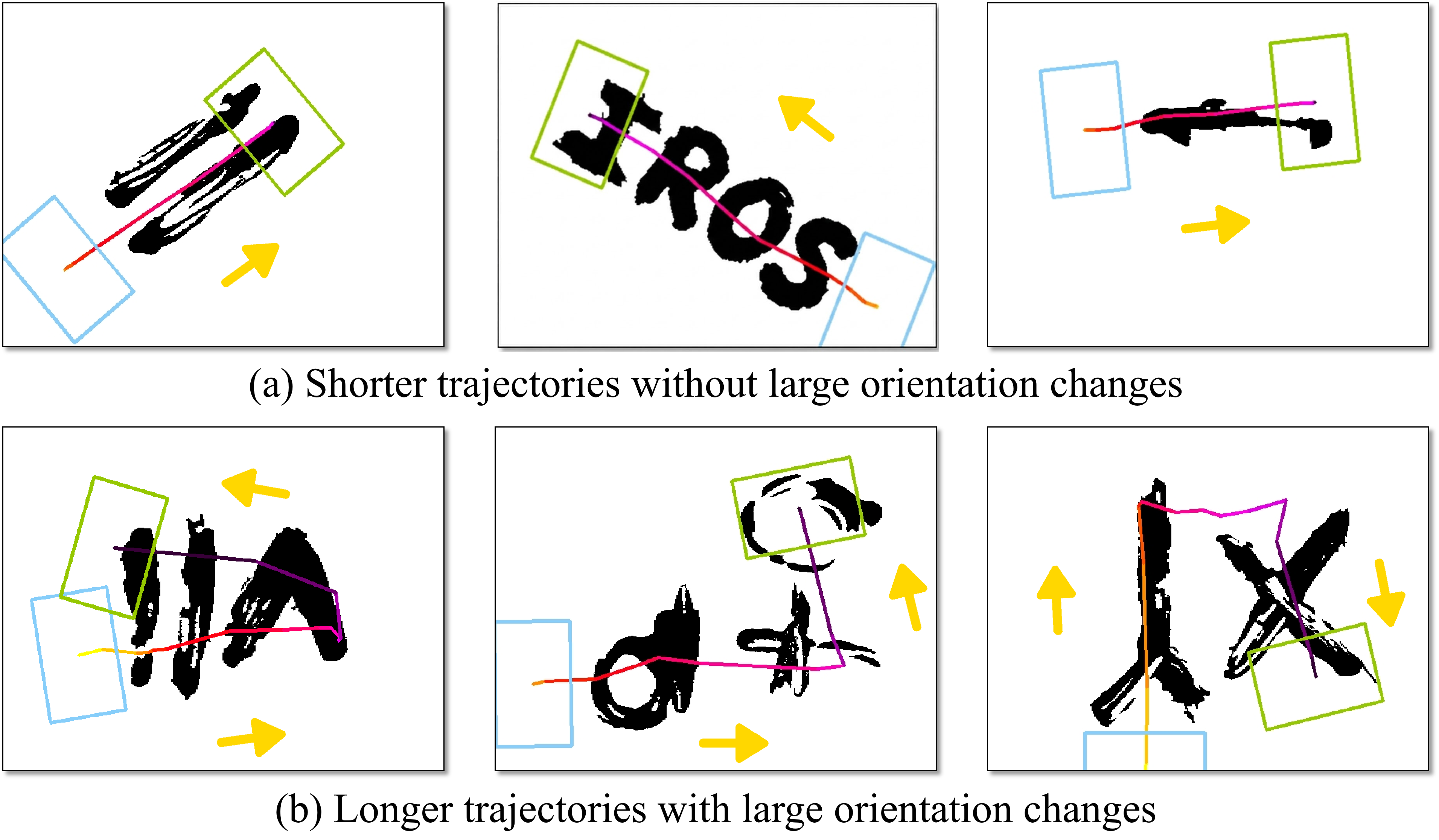}
    \caption{\textbf{PO training trajectories (global sweep plans).} 
    For each stain map, PO plans a single continuous \emph{global} pushing trajectory per iteration to traverse the current stain region, which often yields long paths and large orientation changes. 
    The sponge pose is shown as a blue rectangle at the start and a green rectangle at the end.}
    \label{fig:PO}
\end{figure}

\begin{itemize}

    \item \textbf{Full-Cover (FC):} A fixed full-coverage sweeping trajectory (e.g., arched or boustrophedon-like) that spans the entire cleaning area is executed repeatedly.
    
    \item \textbf{PushAll-Onetime(PO):}
    To test whether our segmented pushing is necessary, we construct a baseline that follows the same episodic \emph{re-observe--plan--execute} loop as Push-Wiper, but uses a fundamentally different \emph{per-iteration plan}.
    At each replanning step, PO predicts a \emph{single global} continuous pushing trajectory intended to traverse and sweep through the entire current stain region in one stroke, rather than producing short strokes that progressively aggregate the stain.
    Consequently, PO typically yields long paths and large orientation changes for large or spatially scattered stains.
    As shown in Fig.~\ref{fig:PO}, we use a lightweight 2D trajectory synthesis tool implemented in Pygame to generate supervisory \emph{global sweep} trajectories on the same binary stain-map domain derived from real-world demonstrations.
    This synthesis is used solely for mask-space \emph{planning supervision} rather than contact-rich execution simulation, enabling a controlled comparison of ``global sweep'' versus our aggregation-first segmented strategy under an identical execution stack.
    
\end{itemize}

\textbf{Data Collection.} We teleoperate the robot with a SpaceMouse to collect expert demonstrations for Push-Wiper. All demonstrations are performed on planar surfaces, including 150 cleaning tasks: 75 with ketchup and 75 with peanut butter, each comprising 25 simple and 50 complex stain patterns. Each task contains multiple segmented pushes, yielding 448 pushing trajectories in total. Since the task is formulated as segmented pushing on a binary map, each trajectory is represented by three DoFs, $a=(x_b,y_b,\Delta\theta)$. This formulation enables straightforward data augmentation by applying affine transformations consistently to both stain maps and trajectories, producing 2,688 trajectories for policy training.

\textbf{Implementation.}
Push-Wiper uses a convolution-based diffusion policy with a DDIM~\cite{song2021ddim} scheduler (sample prediction), trained with 100 diffusion steps and run with 10 steps at inference; we predict $n{=}16$ actions per plan.
A hybrid force--position controller executes at 100\,Hz with $F_z^{des}{=}20$\,N, and planning terminates once the detected stain area drops below 100 pixels.
For fair comparison, all baselines share the same force controller; ASPI is disabled for planar tasks (enabled only for curved-surface generalization); the PO baseline uses identical diffusion settings; and all methods perform one sponge scraping after each executed trajectory.

\begin{table*}[htbp]
\centering
\vspace{1ex}
\captionsetup[table]{labelfont=normalfont,textfont=normalfont,justification=centering,labelsep=none}
\caption{CS of three methods on ketchup and peanut butter (mean$\pm$std over trials).}
\label{tab:main}
\vspace{2mm}

\setlength{\tabcolsep}{3.2pt}
\renewcommand{\arraystretch}{1.05}
\small

\begin{tabularx}{\textwidth}{ 
    >{\centering\arraybackslash}m{3.4cm} 
    >{\centering\arraybackslash}X
    >{\centering\arraybackslash}X
    >{\centering\arraybackslash}X
    >{\centering\arraybackslash}X
    >{\centering\arraybackslash}X
    >{\centering\arraybackslash}X
    >{\centering\arraybackslash}X
}
\toprule
\multirow{2}{*}{\textbf{Method}}
& \multicolumn{3}{c}{\textbf{\textit{Ketchup}}}
& \multicolumn{3}{c}{\textbf{\textit{Peanut butter}}}
& \multirow{2}{*}{\textbf{\textit{Overall Avg}}} \\
\cmidrule(lr){2-4} \cmidrule(lr){5-7}
& \textbf{Simple} & \textbf{Complex} & \textbf{Avg}
& \textbf{Simple} & \textbf{Complex} & \textbf{Avg} &  \\
\midrule
Full-Cover (FC) 
& $54.24\pm20.98$ & $53.77\pm11.99$ & $53.99\pm18.13$
& $16.46\pm31.41$ & $6.11\pm38.45$ & $11.28\pm34.58$ & $32.64\pm34.76$ \\
PushAll-Onetime (PO) 
& $55.33\pm17.88$ & $60.75\pm6.60$ & $58.04\pm13.54$
& $38.41\pm16.13$ & $25.58\pm12.90$ & $31.99\pm15.67$ & $44.98\pm19.44$ \\
\midrule
Push-Wiper (\textit{Ours}) 
& $\mathbf{90.43}\pm4.79$ & $\mathbf{94.18}\pm3.16$ & $\mathbf{92.30}\pm4.73$ 
& $\mathbf{85.79}\pm9.69$ & $\mathbf{89.13}\pm4.88$ & $\mathbf{87.46}\pm7.66$ 
& $\mathbf{89.88}\pm6.80$ \\
\bottomrule
\end{tabularx}
\end{table*}

\begin{figure*}[t]   
    \centering
    \includegraphics[width=0.95\textwidth]{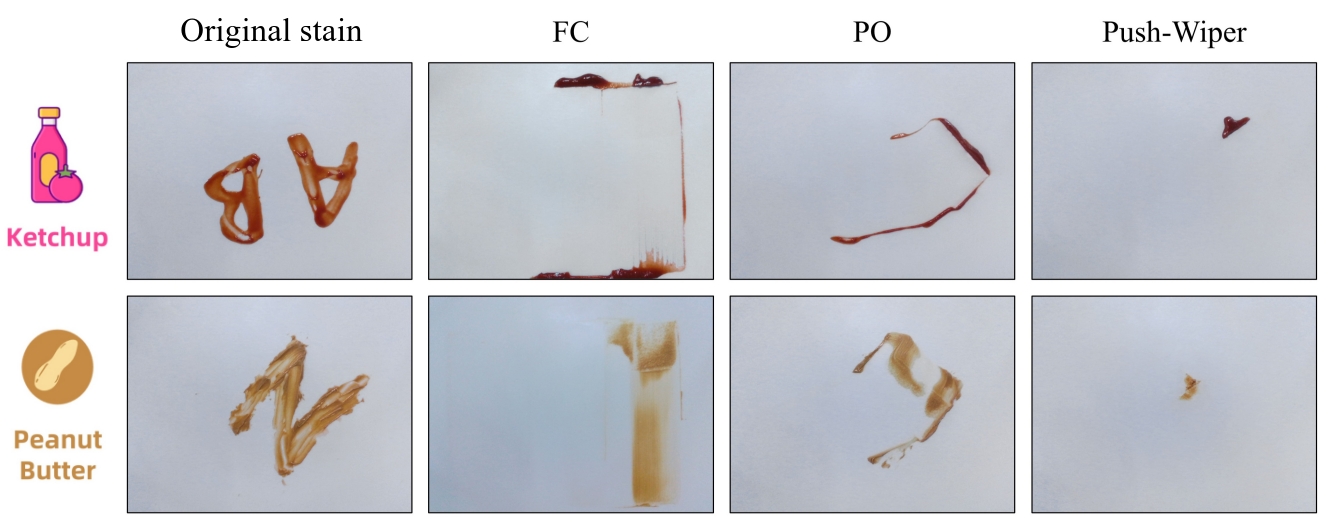}
    \caption{\textbf{Comparison of cleaning results on the same stain distribution using three methods.} Both FC and PO cause varying degrees of secondary contamination, especially on highly viscous peanut butter. In contrast, Push-Wiper consistently achieves superior cleaning performance across all tests.}
    \label{fig:main}
\end{figure*}

\subsection{Cleaning results of three methods}
\begin{figure}[htbp]   
    \centering         
    \includegraphics[width=1.0\linewidth]{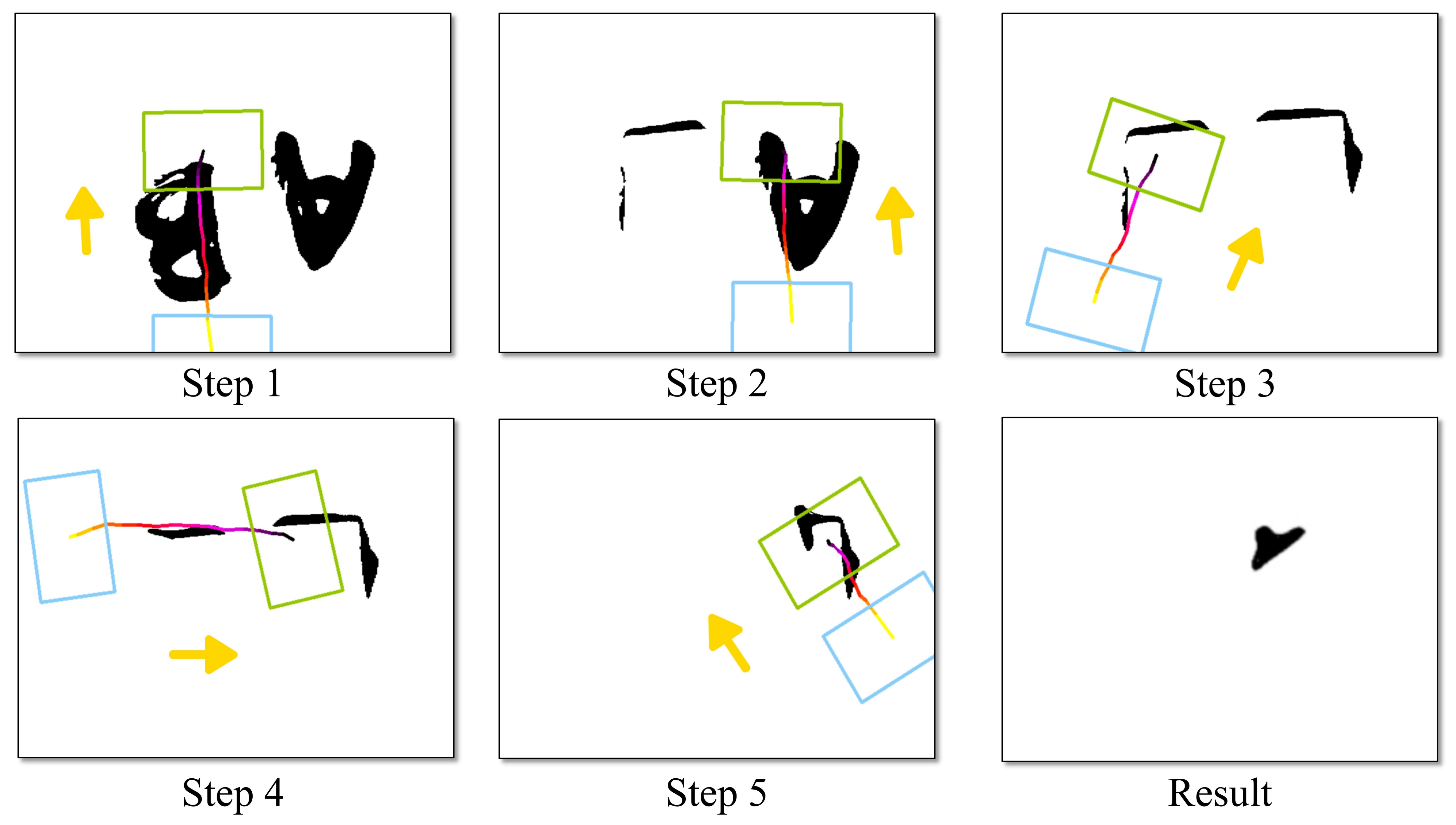}  
    \caption{\textbf{Visualization of a segmented pushing process.} 
    }
    \label{fig:pushstep} 
\end{figure}

For each method, we conduct 20 experiments on ketchup and another 20 on peanut butter, each on distinct stain distributions, with 10 simple and 10 complex cases per stain type. As shown in Table~\ref{tab:main}, Push-Wiper consistently achieves the highest CS and shows strong stability, with an overall average of 89.88, significantly outperforming FC (32.64) and PO (44.98). 
Notably, while FC and PO suffer performance drops on high-viscosity peanut butter compared to ketchup, Push-Wiper maintains similarly high scores across both stain types. 

\renewcommand{\thetable}{\Roman{table}}
\begin{table}[h]
\caption{CS on unseen curved surfaces.}
\centering
\label{tab:curved}
\vspace{2mm}

\setlength{\tabcolsep}{3.2pt}
\renewcommand{\arraystretch}{1.05}
\small

\begin{tabularx}{\linewidth}{
    >{\centering\arraybackslash}m{2.05cm}
    >{\centering\arraybackslash}X
    >{\centering\arraybackslash}X
    >{\centering\arraybackslash}X
    >{\centering\arraybackslash}X
    >{\centering\arraybackslash}X
    >{\centering\arraybackslash}X
}
\toprule
\multirow{2}{*}{\textbf{Method}}
& \multicolumn{3}{c}{\textbf{\textit{Convex}}}
& \multicolumn{3}{c}{\textbf{\textit{Concave}}} \\
\cmidrule(lr){2-4} \cmidrule(lr){5-7}
& \textbf{Ketch.} & \textbf{Pb.} & \textbf{Avg}
& \textbf{Ketch.} & \textbf{Pb.} & \textbf{Avg} \\
\midrule
\makecell{Push-Wiper}
& \makecell{$95.27$\\$\pm0.77$}
& \makecell{$87.63$\\$\pm8.31$}
& \makecell{$91.45$\\$\pm6.86$}
& \makecell{$94.75$\\$\pm2.86$}
& \makecell{$92.10$\\$\pm2.58$}
& \makecell{$93.42$\\$\pm2.93$} \\
\bottomrule
\end{tabularx}
\end{table}

As shown in Fig.~\ref{fig:main}, visual inspection reveals clear failure modes of the baselines.
\textbf{FC} tends to \emph{smear} viscous stains: under force-controlled wiping, sponge deformation drags and redistributes material along the path, causing severe secondary contamination (especially on peanut butter) and leaving substantial residues due to limited absorption, which can even yield negative CS.
\textbf{PO} adapts its sweep to the current stain distribution, yet it lacks an explicit \emph{aggregation} behavior---its global wipe/push trajectories mainly traverse and spread the stain mass rather than progressively consolidating it---so residual fragments persist across iterations and are difficult to eliminate.
In contrast, \textbf{Push-Wiper} executes step-wise pushes that progressively concentrate stains into compact regions, enabling reliable cleanup even under complex and spatially scattered stain distributions.
A segmented pushing process is illustrated in Fig.~\ref{fig:pushstep}.

Besides cleaning effectiveness, Push-Wiper has an average wall-clock runtime of 130\,s per trial, measured from execution start until the stopping criterion (stain area $<100$ pixels) is met; this includes the fixed sponge-scraping step after each pushing trajectory.
As reported in Table~\ref{tab:main}, we use Push-Wiper's completion time as the matched execution budget $T$ for each test case, and run each baseline for as many full trajectories as fit within $T$, with sponge scraping after every trajectory.
Because secondary contamination can make CS non-monotonic with longer runs (Fig.~\ref{fig:main}), we report the best CS each baseline attains at any stopping point within $T$, rather than a time-to-completion metric.
Overall, Push-Wiper achieves strong cleaning performance within a minutes-scale execution budget.

\begin{figure*}[htbp]   
    \centering
    \vspace{1ex}
    \includegraphics[width=\linewidth]{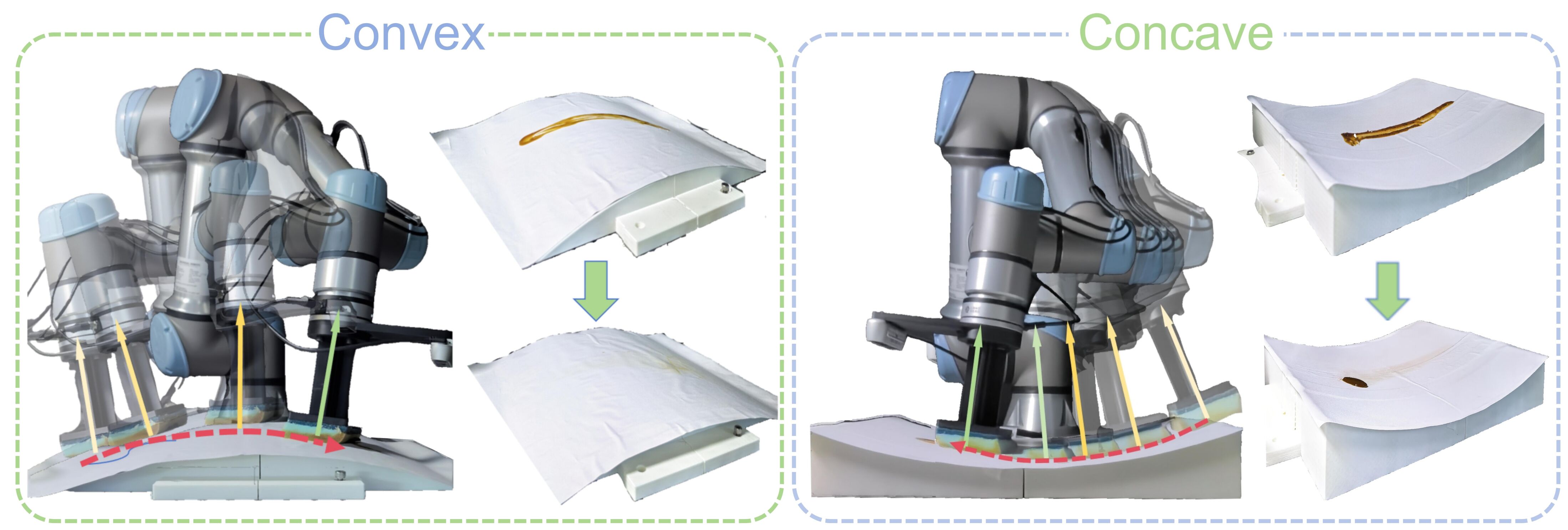}
    \caption{\textbf{Visualization of cleaning tasks on unseen curved surfaces.} On convex and concave surfaces with black pepper sauce and oyster sauce, respectively, both of which are unseen viscous stains.
}
    \label{fig:curve}
\end{figure*}

\begin{figure}[htbp]   
    \centering         
    \includegraphics[width=1\linewidth]{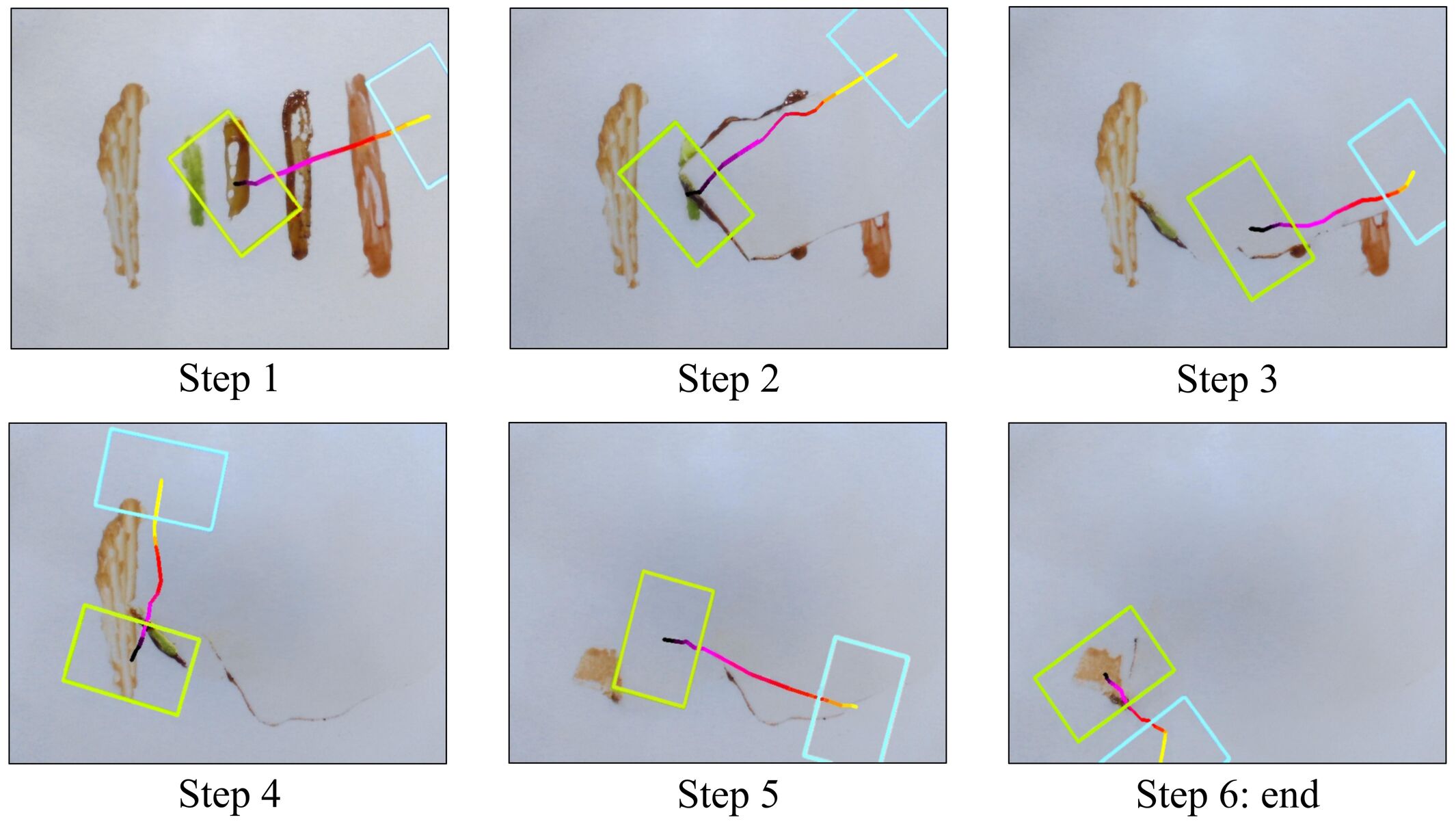}  
    \caption{\textbf{A segmented pushing process and result on CVS.} The sponge block is shown as a blue rectangle at the start and a green rectangle at the end.  
    }
    \label{fig:cvs} 
\end{figure}

\subsection{Generalization on unseen curved surfaces}

We evaluate the generalization of Push-Wiper on curved surfaces using two  geometries: a convex and a concave surface. It should be noted that the training data do not include curved surfaces. For each surface, we perform 10 complex-stain trials (5 ketchup, 5 peanut butter). As shown in Table~\ref{tab:curved}, Push-Wiper achieves CS of 91.45 and 93.42 on the convex and concave surfaces, respectively, comparable to those on planar surfaces. Fig.~\ref{fig:curve} shows that Push-Wiper adaptively generates push trajectories, while ASPI keeps the end-effector normal aligned with the surface normal throughout the motion, thereby demonstrating robust generalization to out-of-distribution surface geometries.

\begin{table}[t]
\caption{CS on unseen objects and stains.}
\centering
\label{tab:push_wiper_comparison}

\setlength{\tabcolsep}{3.2pt}
\renewcommand{\arraystretch}{1.05}
\small

\begin{tabularx}{\linewidth}{ 
    >{\centering\arraybackslash}m{2cm} 
    >{\centering\arraybackslash}X 
    >{\centering\arraybackslash}X 
    >{\centering\arraybackslash}X 
}
\toprule
\textbf{Method} & \textbf{\textit{Solid}} & \textbf{\textit{Liquid}} & \textbf{\textit{CVS}} \\
\midrule
Push-Wiper  & $100.00\pm0.00$ & $92.62\pm7.43$ & $94.42\pm3.80$ \\
\bottomrule
\end{tabularx}
\end{table}

\subsection{Generalization on unseen objects and stains}

\begin{figure}[!htbp]   
    \centering         
    \includegraphics[width=1\linewidth]{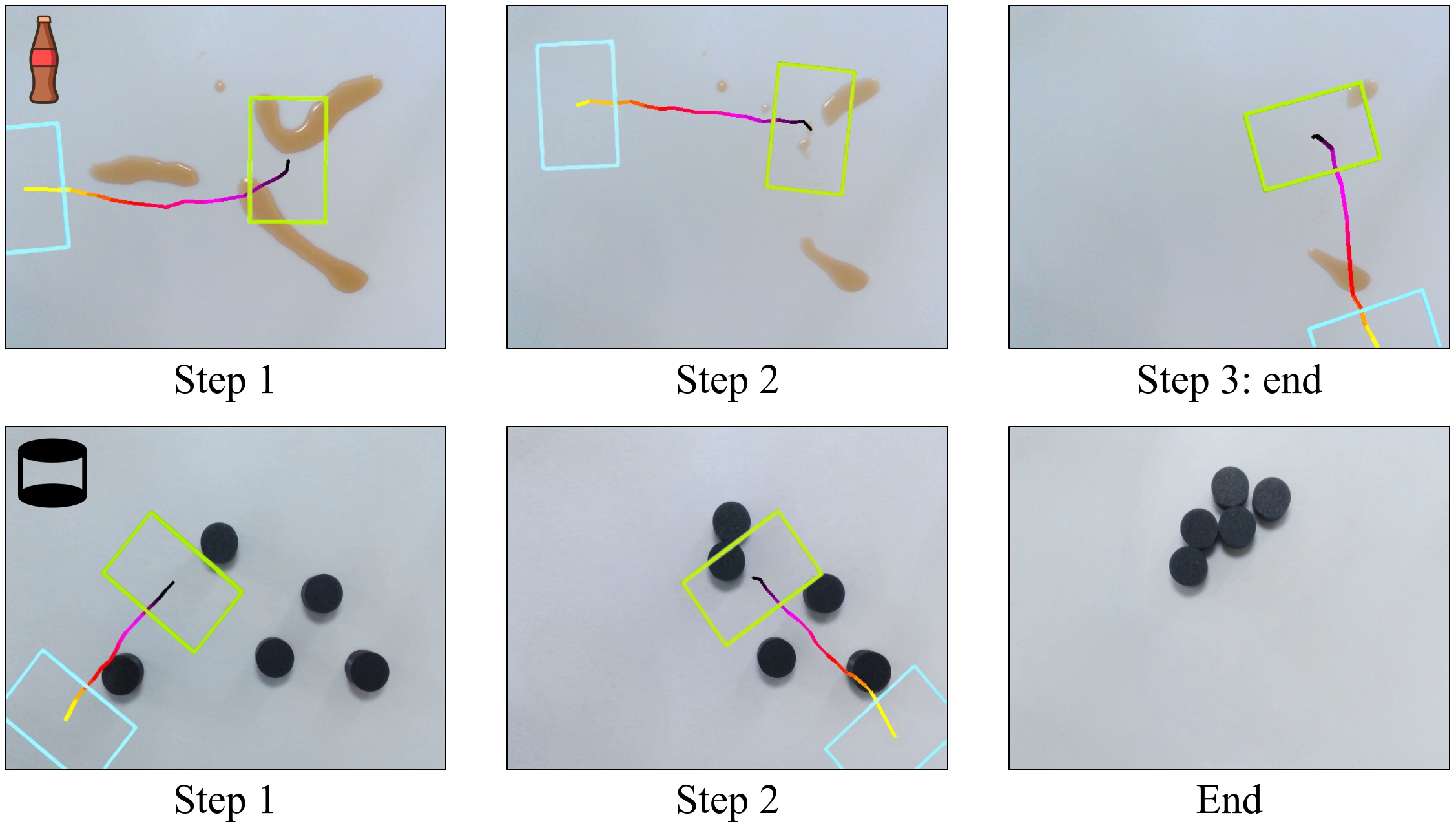}  
    \caption{\textbf{Segmented pushing process and results: top row shows cola, bottom row shows black disks.}  
    }
    \label{fig:solid} 
\end{figure}

We evaluate Push-Wiper on a variety of unseen scenarios, including solids, liquids, and a combination of unseen viscous stains (abbreviated as CVS), such as black pepper sauce and oyster sauce, to assess its zero-shot generalization to other substances. For the black disks, we randomly place five disks on the table. Since solids cannot be removed from the surface using a sponge, we evaluate whether the policy can aggregate them into a small area. If the final number of connected regions is one, the CS is set to 100; otherwise, it is 0. We conduct 10 complex-type experiments for each of solids, liquids, and CVS.
As shown in Table~\ref{tab:push_wiper_comparison}, Fig.~\ref{fig:cvs} and Fig.~\ref{fig:solid}, Push-Wiper not only generalizes to previously unseen viscous stains but also applies to stains of different physical states.

\renewcommand{\thetable}{\Roman{table}} 
\begin{table}[t]
\small
\centering
\captionsetup[table]{labelfont=normalfont,textfont=normalfont,
    justification=centering}
\caption{{w/o. post-processing vs. w. post-processing}}
\label{tab:post_process}

\begin{tabularx}{\linewidth}{ 
    >{\centering\arraybackslash}m{3cm} 
    >{\centering\arraybackslash}X 
    >{\centering\arraybackslash}X 
    >{\centering\arraybackslash}X 
}
\toprule
\textbf{Method} & \textbf{\textit{Ketch.}} & \textbf{\textit{Pb.}} & \textbf{\textit{Avg}} \\
\midrule
\makecell{Push-Wiper \\ \textit{(w/o. post-processing)}} & $93.76$ & $85.12$ & $89.44$ \\
\makecell{Push-Wiper \\ \textit{(w. post-processing)}} 
& $100.00$\textsuperscript{\textcolor{mygreen}{↑\textit{$6.66\%$}}}
& $98.51$\textsuperscript{\textcolor{mygreen}{↑\textit{$15.73\%$}}} 
& $99.25$\textsuperscript{\textcolor{mygreen}{↑\textit{$10.97\%$}}} \\
\bottomrule
\end{tabularx}
\end{table}

\subsection{Evaluation of cleaning effectiveness of post-processing}

The previous experiments show that the gathering phase already achieves near-complete cleaning. We further employ a post-processing phase to enhance performance. Five trials are conducted for ketchup and peanut butter, respectively.
As shown in Table~\ref{tab:post_process}, post-processing increases the CS to 100 for ketchup and 98.51 for peanut butter. These results demonstrate that the combination of the gathering phase with the post-processing enables our framework to achieve virtually complete cleaning.

\section{Conclusions}

In this paper, we present Push-Wiper, a framework for cleaning viscous stains from arbitrary surfaces. Push-Wiper uses a sequence of segmented pushing trajectories to aggregate viscous material and thereby mitigate secondary contamination. A stepwise policy governed by Diffusion Policy adapts to diverse spatial distributions of stains, and the generated actions are executed via our ASPI together with a hybrid force–position controller. In comparative experiments on ketchup and peanut butter, Push-Wiper outperforms baseline methods, achieving an average cleaning score improvement of approximately 130\%. We compare results before and after post-processing to verify its effectiveness. Push-Wiper also generalizes to diverse surface geometries and handles solid residues, liquid spills, and unseen viscous stains without retraining. Future work will add collision constraints for safe human–robot interaction and extend post-processing to a broader range of stain types.

\addtolength{\textheight}{-12cm}   





\bibliographystyle{IEEEtran}
\bibliography{reference}

@article{megalingam2025cleaning,
  title={Cleaning robots: A review of sensor technologies and intelligent control strategies for cleaning},
  author={Megalingam, Rajesh Kannan and Vadivel, Shree Rajesh Raagul and Kotaprolu, Sai Smaran and Nithul, Bagathi and Kumar, Devisetty Vijay and Rudravaram, Gaurav},
  journal={Journal of Field Robotics},
  year={2025},
  publisher={Wiley Online Library}
}

@inproceedings{bordoloi2017floor,
  title={A floor cleaning robot for domestic environments},
  author={Bordoloi, Arnab K and Islam, Md Faheemul and Zaman, Jiauz and Phukan, Nabasmita and Kakoty, Nayan M},
  booktitle={Proceedings of the 2017 3rd International Conference on Advances in Robotics},
  pages={1--5},
  year={2017}
}

@article{li2021survey,
  title={A survey on techniques and applications of window-cleaning robots},
  author={Li, Zhenjing and Xu, Qingsong and Tam, Lap Mou},
  journal={IEEE Access},
  volume={9},
  pages={111518--111532},
  year={2021},
  publisher={IEEE}
}

@inproceedings{harmatz2024hybrid,
  title={Hybrid force-position control of an elastic tendon-driven scrubbing robot (tedsr)},
  author={Harmatz, Noah and Zahra, Alina and Abdelmalak, Amir and Purohit, Shivam and Shin, Trevor and Mazzeo, Aaron D},
  booktitle={2024 IEEE International Conference on Robotics and Automation (ICRA)},
  pages={4693--4699},
  year={2024},
  organization={IEEE}
}

@article{kowalewski2025sccrub,
  title={SCCRUB: Surface Cleaning Compliant Robot Utilizing Bristles},
  author={Kowalewski, Jakub F and Hajjafar, Keeyon and Ugent, Alyssa and Lipton, Jeffrey Ian},
  journal={arXiv preprint arXiv:2507.06053},
  year={2025}
}

@article{landel2021fluid,
  title={The fluid mechanics of cleaning and decontamination of surfaces},
  author={Landel, Julien R and Wilson, D Ian},
  journal={Annual Review of Fluid Mechanics},
  volume={53},
  number={1},
  pages={147--171},
  year={2021},
  publisher={Annual Reviews}
}

@article{kim2019control,
  title={Control strategies for cleaning robots in domestic applications: A comprehensive review},
  author={Kim, Jaeseok and Mishra, Anand Kumar and Limosani, Raffaele and Scafuro, Marco and Cauli, Nino and Santos-Victor, Jose and Mazzolai, Barbara and Cavallo, Filippo},
  journal={International Journal of Advanced Robotic Systems},
  volume={16},
  number={4},
  pages={1729881419857432},
  year={2019},
  publisher={SAGE Publications Sage UK: London, England}
}

@article{wakabayashi2024behavioral,
  title={Behavioral learning of dish rinsing and scrubbing based on interruptive direct teaching considering assistance rate},
  author={Wakabayashi, Shumpei and Kawaharazuka, Kento and Okada, Kei and Inaba, Masayuki},
  journal={Advanced Robotics},
  volume={38},
  number={15},
  pages={1052--1065},
  year={2024},
  publisher={Taylor \& Francis}
}

@book{johansson2007handbook,
  title={Handbook for cleaning/decontamination of surfaces},
  author={Johansson, Ingegard and Somasundaran, Ponisseril},
  year={2007},
  publisher={Elsevier}
}

@article{yin2020table,
  title={Table cleaning task by human support robot using deep learning technique},
  author={Yin, Jia and Apuroop, Koppaka Ganesh Sai and Tamilselvam, Yokhesh Krishnasamy and Mohan, Rajesh Elara and Ramalingam, Balakrishnan and Le, Anh Vu},
  journal={Sensors},
  volume={20},
  number={6},
  pages={1698},
  year={2020},
  publisher={MDPI}
}

@INPROCEEDINGS{Lew2023,
  author={Lew, Thomas and Singh, Sumeet and Prats, Mario and Bingham, Jeffrey and Weisz, Jonathan and Holson, Benjie and Zhang, Xiaohan and Sindhwani, Vikas and Lu, Yao and Xia, Fei and Xu, Peng and Zhang, Tingnan and Tan, Jie and Gonzalez, Montserrat},
  booktitle={2023 IEEE International Conference on Robotics and Automation (ICRA)}, 
  title={Robotic Table Wiping via Reinforcement Learning and Whole-body Trajectory Optimization}, 
  year={2023},
  volume={},
  number={},
  pages={7184-7190},
  doi={10.1109/ICRA48891.2023.10161283}}

@article{chi2023diffusion,
  title={Diffusion policy: Visuomotor policy learning via action diffusion},
  author={Chi, Cheng and Xu, Zhenjia and Feng, Siyuan and Cousineau, Eric and Du, Yilun and Burchfiel, Benjamin and Tedrake, Russ and Song, Shuran},
  journal={The International Journal of Robotics Research},
  pages={02783649241273668},
  year={2023},
  publisher={SAGE Publications Sage UK: London, England}
}

@inproceedings{hess2012null,
  title={Null space optimization for effective coverage of 3d surfaces using redundant manipulators},
  author={Hess, J{\"u}rgen and Tipaldi, Gian Diego and Burgard, Wolfram},
  booktitle={2012 IEEE/RSJ International Conference on Intelligent Robots and Systems},
  pages={1923--1928},
  year={2012},
  organization={IEEE}
}

@inproceedings{Wang2025Hierarchically,
  author={Wang, Yeping and Gleicher, Michael},
  booktitle={2025 IEEE International Conference on Robotics and Automation (ICRA)}, 
  title={Hierarchically Accelerated Coverage Path Planning for Redundant Manipulators}, 
  year={2025},
  volume={},
  number={},
  pages={12098-12104},
  doi={10.1109/ICRA55743.2025.11128545}}

@inproceedings{ortenzi2014experimental,
  title={An experimental study of robot control during environmental contacts based on projected operational space dynamics},
  author={Ortenzi, Valerio and Adjigble, Maxime and Kuo, Jeffrey A and Stolkin, Rustam and Mistry, Michael},
  booktitle={2014 IEEE-RAS International Conference on Humanoid Robots},
  pages={407--412},
  year={2014},
  organization={IEEE}
}

@article{leidner2016knowledge,
  title={Knowledge-enabled parameterization of whole-body control strategies for compliant service robots},
  author={Leidner, Daniel and Dietrich, Alexander and Beetz, Michael and Albu-Sch{\"a}ffer, Alin},
  journal={Autonomous Robots},
  volume={40},
  number={3},
  pages={519--536},
  year={2016},
  publisher={Springer}
}

@inproceedings{hogan1984impedance,
  title={Impedance control: An approach to manipulation},
  author={Hogan, Neville},
  booktitle={1984 American control conference},
  pages={304--313},
  year={1984},
  organization={IEEE}
}

@inproceedings{martin2019variable,
  title={Variable impedance control in end-effector space: An action space for reinforcement learning in contact-rich tasks},
  author={Mart{\'\i}n-Mart{\'\i}n, Roberto and Lee, Michelle A and Gardner, Rachel and Savarese, Silvio and Bohg, Jeannette and Garg, Animesh},
  booktitle={2019 IEEE/RSJ international conference on intelligent robots and systems (IROS)},
  pages={1010--1017},
  year={2019},
  organization={IEEE}
}

@article{tsuji2024adaptive,
  title={Adaptive contact-rich manipulation through few-shot imitation learning with Force-Torque feedback and pre-trained object representations},
  author={Tsuji, Chikaha and Coronado, Enrique and Osorio, Pablo and Venture, Gentiane},
  journal={IEEE Robotics and Automation Letters},
  year={2024},
  publisher={IEEE}
}

@article{oishi2025imitation,
  title={Imitation Learning Based on Disentangled Representation Learning of Behavioral Characteristics},
  author={Oishi, Ryoga and Sakaino, Sho and Tsuji, Toshiaki},
  journal={arXiv preprint arXiv:2509.04737},
  year={2025}
}

@article{kumbar2019rheological,
  title={Rheological properties of tomato ketchup.},
  author={Kumb{\'a}r, Vojt{\v{e}}ch and Ondru{\v{s}}{\'\i}kov{\'a}, Sylvie and Nedomov{\'a}, {\v{S}}{\'a}rka},
  year={2019}
}

@article{citerne2001rheological,
  title={Rheological properties of peanut butter},
  author={Citerne, Guillaume P and Carreau, Pierre J and Moan, Michel},
  journal={Rheologica Acta},
  volume={40},
  number={1},
  pages={86--96},
  year={2001},
  publisher={Springer}
}

@inproceedings{ott2010unified,
  title={Unified impedance and admittance control},
  author={Ott, Christian and Mukherjee, Ranjan and Nakamura, Yoshihiko},
  booktitle={2010 IEEE international conference on robotics and automation},
  pages={554--561},
  year={2010},
  organization={IEEE}
}

@inproceedings{song2021ddim,
  title={Denoising Diffusion Implicit Models},
  author={Song, Jiaming and Meng, Chenlin and Ermon, Stefano},
  booktitle={International Conference on Learning Representations (ICLR)},
  year={2021}
}

@article{zhou2025admittance,
  title={Admittance visuomotor policy learning for general-purpose contact-rich manipulations},
  author={Zhou, Bo and Jiao, Ruixuan and Li, Yi and Yuan, Xiaogang and Fang, Fang and Li, Shihua},
  journal={IEEE Transactions on Industrial Electronics},
  year={2025},
  publisher={IEEE}
}

@article{he2025foar,
  title={FoAR: Force-Aware Reactive Policy for Contact-Rich Robotic Manipulation},
  author={He, Zihao and Fang, Hongjie and Chen, Jingjing and Fang, Hao-Shu and Lu, Cewu},
  journal={IEEE Robotics and Automation Letters},
  year={2025},
  publisher={IEEE}
}

@inproceedings{xue2025reactive,
  title     = {Reactive Diffusion Policy: Slow-Fast Visual-Tactile Policy Learning for Contact-Rich Manipulation},
  author    = {Xue, Han and Ren, Jieji and Chen, Wendi and Zhang, Gu and Fang, Yuan and Gu, Guoying and Xu, Huazhe and Lu, Cewu},
  booktitle = {Proceedings of Robotics: Science and Systems (RSS)},
  year      = {2025}
}

@article{song2025survey,
  title={A survey on diffusion policy for robotic manipulation: Taxonomy, analysis, and future directions},
  author={Song, Mingchen and Deng, Xiang and Zhou, Zhiling and Wei, Jie and Guan, Weili and Nie, Liqiang},
  journal={Authorea Preprints},
  year={2025},
  publisher={Authorea}
}

@inproceedings{xian2023fluidlab,
  title={FluidLab: A Differentiable Environment for Benchmarking Complex Fluid Manipulation},
  author={Xian, Zhou and Zhu, Bo and Xu, Zhenjia and Tung, Hsiao-Yu and Torralba, Antonio and Fragkiadaki, Katerina and Gan, Chuang},
  booktitle={International Conference on Learning Representations},
  year={2023}
}

@article{billard2019trends,
  title={Trends and challenges in robot manipulation},
  author={Billard, Aude and Kragic, Danica},
  journal={Science},
  volume={364},
  number={6446},
  pages={eaat8414},
  year={2019},
  publisher={American Association for the Advancement of Science}
}

@INPROCEEDINGS{2019Song,
  author={Song, Daeun and Kim, Young J.},
  booktitle={2019 International Conference on Robotics and Automation (ICRA)}, 
  title={Distortion-free Robotic Surface-drawing using Conformal Mapping}, 
  year={2019},
  volume={},
  number={},
  pages={627-633},
  doi={10.1109/ICRA.2019.8794034}}

\end{document}